%% file: conference_101719.tex
\documentclass[conference]{IEEEtran}
\IEEEoverridecommandlockouts
\usepackage{amsmath,amssymb,amsfonts}
\usepackage{algorithmic}
\usepackage{graphicx}
\usepackage{textcomp}
\usepackage{orcidlink}
\usepackage{xcolor}
\usepackage{tikz}
\usetikzlibrary{plotmarks}              % larger choice of plot marks
\usetikzlibrary{arrows}                 % larger choice of arrow heads
\usetikzlibrary{arrows.meta, positioning, calc}
\usepackage{pgfplots}
\pgfplotsset{compat=1.18}
\usepgfplotslibrary{groupplots}
\usepackage{subcaption}
\def\BibTeX{{\rm B\kern-.05em{\sc i\kern-.025em b}\kern-.08em
    T\kern-.1667em\lower.7ex\hbox{E}\kern-.125emX}}
\begin{document}

\title{Exploring AI-based Anonymization of Industrial Image and Video Data in the Context of Feature Preservation}

\author{
\IEEEauthorblockN{Sabrina C. Triess \orcidlink{0009-0003-0839-901X}}
\IEEEauthorblockA{\textit{Divison Image and Signal Processing} \\
\textit{Fraunhofer IPA}\\
Stuttgart, Germany}
\and
\IEEEauthorblockN{Timo Leitritz \orcidlink{0009-0009-2109-4724}}
\IEEEauthorblockA{\textit{Divison Image and Signal Processing} \\
\textit{Fraunhofer IPA}\\
Stuttgart, Germany}
\and
\IEEEauthorblockN{Christian Jauch \orcidlink{0000-0002-2769-2831}}
\IEEEauthorblockA{\textit{Division Image and Signal Processing} \\
\textit{Fraunhofer IPA}\\
Stuttgart, Germany}
}

\maketitle

\hyphenation{dri-ving qua-li-ty ano-ny-mi-zing wor-king eva-lu-a-tion un-re-cog-ni-zable vi-sible thres-hold labo-ra-to-ry using shoo-ting cha-rac-te-ris-tics ma-nu-fac-tu-ring}

%%%%%%%%%%%%%%%%%%%%%%%%%%%%%%%%%%%%%%%%%%%%%%%%%%%%%%%%%%%%%%%%%%%%%%%%%%%%%%%%
\begin{abstract}
With rising technologies, the protection of privacy-sensitive information is becoming increasingly important. 
In industry and production facilities, image or video recordings are beneficial for documentation, tracing production errors or coordinating workflows. 
Individuals in images or videos need to be anonymized. 
However, the anonymized data should be reusable for further applications. 
In this work, we apply the Deep Learning-based full-body anonymization framework DeepPrivacy2, which generates artificial identities, to industrial image and video data. 
We compare its performance with conventional anonymization techniques. 
Therefore, we consider the quality of identity generation, temporal consistency, and the applicability of pose estimation and action recognition.
\end{abstract}

%%%%%%%%%%%%%%%%%%%%%%%%%%%%%%%%%%%%%%%%%%%%%%%%%%%%%%%%%%%%%%%%%%%%%%%%%%%%%%%%


%%%%%%%%%%%%%%%%%%%%%%%%%%%%%%%%%%%%%%%%%%%%%%%%%%%%%%%%%%%%%%%%%%%%%%%%%%%%%%%%
\section{INTRODUCTION}

With rising technologies like human activity recognition and object detection in various applications like autonomous driving and smart homes, the protection of privacy-sensitive data is becoming increasingly important. 
Images and videos showing people who are identifiable based on their appearance, whether physical features such as face and body shape or characteristics such as clothing and haircut, must be anonymized. 

In this work, we consider the recording of industrial manufacturing processes performed by humans, where even the tiniest characteristics can be sufficient for identification due to the limited number of people within a work group or department.
Such processes can be sorting of work components, operating machines, assembling or just monitoring for the safety of employees in production facilities. 
The goal is to anonymize a person in such a way that he or she cannot be identified by a recognition system or by other persons known or unknown to that person. 
It should not be possible to draw conclusions about the identity either inside or outside the working environment. 
Nevertheless, the data should remain usable for face or body detectors and further applications. 
The generated artificial identities should look realistic to the human eye, conform to the pose, and be temporally consistent.
DeepPrivacy2, a GAN-based detection and full-body anonymization architecture by Hukkelås \emph{et al.}~\cite{hukkelas23DP2}, promises to meet all these requirements.\\

The contributions of this work are: 
\begin{itemize}
    \item Evaluation of DeepPrivacy2 in an industrial setting
    \item Analysis of pose estimation and action recognition on anonymized images and videos
    \item Comparison with the conventional anonymization techniques blurring and pixelation
\end{itemize}

%%%%%%%%%%%%%%%%%%%%%%%%%%%%%%%%%%%%%%%%%%%%%%%%%%%%%%%%%%%%%%%%%%%%%%%%%%%%%%%%
\section{RELATED WORK}

Naive anonymization methods such as pixelation, blurring or masking cover privacy-sensitive regions of individuals, but they cause data loss and degradation of image quality. 
The consequence is unusable data such that detection algorithms or downstream tasks might fail. 
Most state-of-the-art anonymization approaches based on neural networks focus only on face anonymization. 
Features outside the face, such as haircut, tattoos, clothing, or other special characteristics, can also reveal an identity and are usually not assessed. 
Korshunova \emph{et al.}~\cite{korshunova2017fast}~implement a Convolutional Neural Network (CNN) that replaces the face in an image with an existing identity. 
Although the actual identity is substituted, this method leads to privacy issues related to the source identity. 

Recent work focuses on Generative Adversarial Networks (GANs) with the goal of generating realistic faces with artificial identities. 
Hukkelås \emph{et al.}~\cite{hukkelås2019deepprivacy} introduce DeepPrivacy, a conditional GAN that removes privacy-sensitive information and generates artificial faces. 
DeepPrivacy only works for images and misses the temporal consistency required for videos. 
Furthermore, the generation of realistic-looking faces fails for irregular poses, profile views, or occluded faces. 
The approach of Gafni \emph{et al.}~\cite{gafni2019live}~focuses on anonymizing videos and alters faces to fool recognition systems. 
However, the generated identities are fixed for a given face and can still be recognized by humans. 
To overcome these issues, Maximov \emph{et al.}~\cite{Maximov_2020}~propose CIAGAN, a conditional GAN for face and full-body anonymization applicable on images and videos. 
With the use of face landmarks and an identity control vector, this approach promises to achieve pose preservation, temporal consistency, and a variety of controllable identities. 
Some generated faces look unnatural to humans and the resolution is limited to $128\times128$. 
As with all approaches that use landmarks, e.g., for pose preservation, landmark detection may fail or reveal privacy-sensitive information. 
In both cases, anonymization is not guaranteed. 
Balaji \emph{et al.}~\cite{balaji2021temporally} apply GAN inpainting for temporally coherent video anonymization without using facial landmarks. 
This approach focuses on temporal consistency between video frames rather than improving anonymization. 
Ma \emph{et al.} present CFA-Net~\cite{ma2021cfanet}~which works for images and videos without the use of landmarks. 
Similar to previous approaches, face-independent features are not considered and the generated faces resemble the original faces. 
%***Alternative (zu lang): Similar to previous approaches, face-independent features are not considered and the generated faces have a high similarity to the original faces.***

In this work, DeepPrivacy2 (DP2), a detection and full-body anonymization architecture by Hukkelås \emph{et al.}~\cite{hukkelas23DP2} is evaluated. 
It is built on DeepPrivacy and on Surface-Guided GANs~\cite{hukkelås2022realistic}. 
No landmark detection is used in this approach, so no privacy-sensitive information is processed. 
Instead, three detection networks are used to ensure detection of all individuals in an image or video: Dual Shot Face Detector (DSFD)~\cite{li2019dsfd} for face detection, segmentation masks from Continuous Surface Embeddings (CSE)~\cite{neverova2020continuous} that do not include hair and clothing on the human body, and segmentation masks from Mask R-CNN~\cite{he2018mask} that include hair and clothing. 
The framework combines three independently trained generators. A face generator, a full-body generator that concatenates CSE to the input and a full-body generator that does not concatenate CSE to the input. 
The generator is based on the U-Net architecture~\cite{ronneberger2015unet} and StyleGAN2~\cite{karras2020analyzing}. 
Both full-body generators are trained on the Flickr Diverse Humans (FDH) dataset~\cite{hukkelas23DP2}. 
The face generator is trained on a higher resolution variant of the Flickr Diverse Faces (FDF) dataset~\cite{hukkelås2019deepprivacy}, FDF256. 

%\addtolength{\textheight}{-3cm}   % This command serves to balance the column lengths

%%%%%%%%%%%%%%%%%%%%%%%%%%%%%%%%%%%%%%%%%%%%%%%%%%%%%%%%%%%%%%%%%%%%%%%%%%%%%%%%
\section{EVALUATION METHODS}
\begin{figure*}[h!]
    \centering
    \scalebox{0.14}{\includegraphics{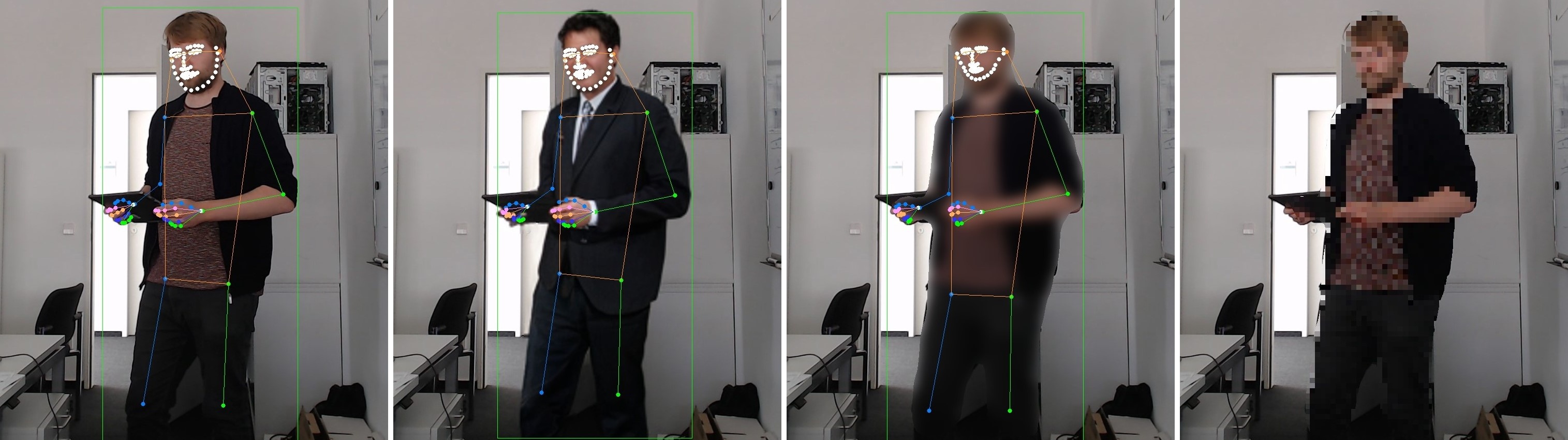}}
    \caption[Pose estimation on RGB images]{Pose estimation on original, DeepPrivacy2-anonymized, blurred and pixelated images (from left to right). The "Office" scenario is shown.}
    \label{fig:pose_estimation_rgb}
\end{figure*}

\subsection{Anonymization}
For anonymization, the DeepPrivacy2 repository~\cite{hukkelas23DP2} is used with the provided checkpoints and default values. 
Full-body anonymization is performed for both images and videos. 
These recordings are made in the office, in a laboratory and in an industrial company to show a realistic industrial working environment with appropriate work processes. 
The evaluation includes anonymization quality for different shooting angles, temporal consistency, and the applicability of whole-body detectors for pose estimation and action recognition. 
For comparison, we apply conventional anonymization techniques, such as blurring and pixelation, to the original data. 
For this purpose, a Mask R-CNN~\cite{he2018mask} generates the segmentation masks of the individuals, which are later anonymized. 
The mask is blurred using a Gaussian filter. 
Pixelation is achieved by reducing the input to the size of the desired pixelation and then enlarging it to the original input size. 
The degree of blurring or pixelation can be set before inference. 

\subsection{Pose Estimation}
The performance of a body detector for 2D human pose estimation is investigated on four image and video types: original, DeepPrivacy2-anonymized, blurred, and pixelated. 
The applied model is provided by the open-source toolbox MMPose~\cite{mmpose2020} and has a High-Resolution Network~\cite{wang2020deep} as its backbone. 
It is pre-trained on the original COCO dataset~\cite{lin2015microsoft} and fine-tuned on the COCO-WholeBody dataset~\cite{jin2020wholebody}. 
A Mask R-CNN is applied for person detection.
A total of 133 keypoints are estimated, each with a certain score. 
Only keypoints with a score greater than a specific threshold are considered detected. 
The remaining coordinates are discarded. 
For evaluation, the commonly used distance metric \emph{End-Point Error (EPE)}~\cite{zimmermann2017learning} is calculated. 
The EPE is the average euclidean distance 

\begin{equation}
\label{equ:error}
    EPE = \frac{1}{N} \cdot \sum_{i = 1}^{N}{\left\|\textbf{x}_{pred,i} - \textbf{x}_{true,i}\right\|_2}
\end{equation}

between the predicted keypoints~$\textbf{x}_{pred}$ and the ground truth~$\textbf{x}_{true}$ over all keypoints $N$. 

Since the goal is to analyze the quality of the anonymized data, it is compared to the original data. 
That means, the detected keypoints on the original image or video frame are treated as ground truth. 
The keypoints of each anonymized image or video are compared to the corresponding ground truth (original keypoint). 

Additionally, we introduce the metric~\emph{Percentage of Total Keypoints (PTK)}, which determines the number of keypoints detected in the anonymized data in relation to the total number of keypoints detected in the original data. 
A PTK of~100\% means, that no keypoints are lost due to anonymization. 

\subsection{Action Recognition}
Similar to pose estimation, we apply human action recognition to original, DeepPrivacy2-anonymized, blurred, and pixelated video recordings. 
The model is taken from MMAction2~\cite{2020mmaction2} and performs spatio-temporal action detection. 
For person detection and action classification, Faster R-CNN~\cite{ren2015faster} trained on the COCO dataset, and SlowOnly-ResNet-101 trained on the AVA dataset~\cite{gu2018ava} are used, respectively. 
We use a label map of 60 action classes to annotate the detected actions within a video frame. 
Each action class is predicted with a certain score. 
Only actions whose score is above a specific threshold are considered detected. 
Actions that are below this threshold are discarded. 
This means, the lower the score, the more actions are considered detected in one frame. 
The class IDs~$x_j \in \mathbb{X}_i$ of the detected actions are stored in a set~$\mathbb{X}_i$ per frame~$i$. 
If no actions are recognized, the set is empty: $\mathbb{X}_i = \emptyset$. 
We treat the detected actions in the original video as the ground truth~$\mathbb{X}_{true}$ and compare the actions detected in the anonymized videos~$\mathbb{X}_{pred}$ to this ground truth. 
For analysis, the action \emph{accuracy (ACC)} is calculated. 
Given a specific score, the anonymized frames are counted in which actions are correctly detected, i.e., are also detected in the original frame. 
This number is divided by the total number of analyzed frames $N$
\begin{equation}
\label{equ:accuracy}
    ACC = \frac{1}{N} \cdot \sum_{i = 1}^{N}{\delta \left(\mathbb{X}_{pred,i} = \mathbb{X}_{true,i}\right)} ,
\end{equation}

where $\delta$ is the indicator function that takes the value~1~if its argument is true. 
The metric~\emph{Percentage of Total Actions (PTA)} counts the anonymized frames with at least one detected action regardless of their correctness and divides them by the number of original frames with at least one detected action. 
Thus, frames without any actions are not taken into account. 
A target value of~100\% shows that the recognition quality is not affected by anonymization. 

%%%%%%%%%%%%%%%%%%%%%%%%%%%%%%%%%%%%%%%%%%%%%%%%%%%%%%%%%%%%%%%%%%%%%%%%%%%%%%%%
\section{EXPERIMENTAL RESULTS}
Full-body anonymization by DeepPrivacy2 is analyzed using an image dataset and a video dataset. 
The image dataset contains 38 images from 4 different scenarios with varying shooting angles: Office (RGB, horizontal), Visionlab assembly (RGB, horizontal), Visionlab machine (grayscale, steep from above), Industrial machine (grayscale, steep from above). 
The video dataset contains 14 video recordings from different shooting angles. 
That is a total of 21,111 frames, and in 17,986 of them, actions are detected when the score is neglected, i.e., the score threshold is set to~0.0. 

\subsection{Visual Quality of DeepPrivacy2}
For horizontal shooting angles, DeepPrivacy2 generates realistic-looking identities with high similarity within a scenario. 
While the posture of the face is well preserved, the facial expression changes. 
The clothes of a person are anonymized regardless of their style. 
For example, the length of the sleeves, the color or the combination of shirt and jacket is not preserved. 
Thus, a person in a production facility is not recognizable by his or her work clothes. 

Furthermore, various visual artifacts can be observed. 
Bent or closed hands in the original image look unrealistic in the anonymized image, and anonymized faces from a steep shooting angle look unnatural. 
Visual artifacts on the hands obscure work components and make work processes unrecognizable as the worktop between body and hands becomes blurred. 
This is not a problem if the components are large and actions such as sorting are performed, as no precision work is done. 
In precision work such as screwing, it becomes unrecognizable what the hands are doing, and the work processes can no longer be traced with the human eye. 

\begin{figure}[h!]
    \centering
    \scalebox{0.721}{\input{plots/IPA_images_EPE_score_50}}
    \caption[End-Point Error of images]{End-Point Error per scenario and anonymization type at a score threshold of~0.5 with respect to~PTK. Each marker represents one image with at least one detected keypoint.}
    \label{fig:epe_images}
\end{figure}
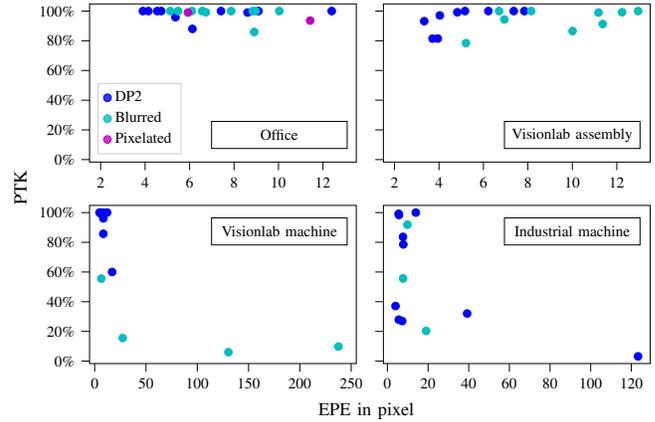

In videos, face anonymization leads to fluctuations between anonymized and original frames due to head movement. 
In contrast, full-body anonymization results in a constant anonymization throughout an entire video and identity is not revealed. 
Variations in identity and clothing can be observed from frame to frame, which distracts from the actual video content. 

\subsection{Pose Estimation on Anonymized Images}
The metrics PTK and EPE for pose estimation are calculated over the image dataset. 
PTK shows how many of the original keypoints are still present in the anonymized data, and EPE shows their deviation. 
Fig.~\ref{fig:epe_images} shows one subplot per scenario and each marker represents one anonymized image in which at least one keypoint is successfully detected. 
Only keypoints with scores greater than 0.5 on both the original and the anonymized data are compared. 

\begin{figure}[h!]
    \centering
    \scalebox{0.95}{\input{plots/IPA_videos_action_distribution}}
    \caption[Action distribution]{Action distribution of~21,111 video frames in total at a score threshold of~0.2. Only the top five actions are shown.}
    \label{fig:action_dist}
\end{figure}
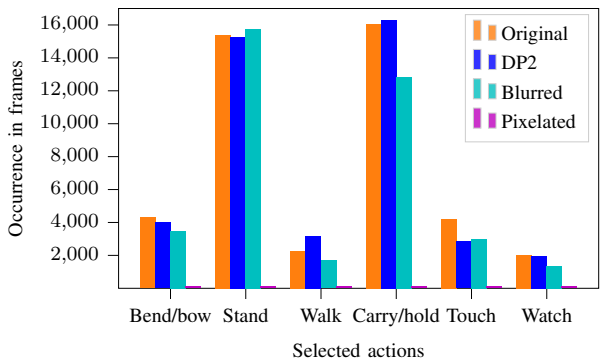

\begin{figure}[h!]
    \centering
    \scalebox{0.95}{\input{plots/IPA_videos_PTA}}
    \caption[Percentage of Total Actions]{Percentage of Total Actions per anonymization type of~21,111 video frames in total.}
    \label{fig:pta}
\end{figure}
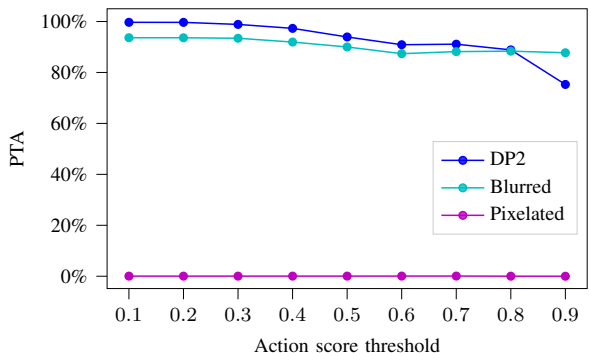

The number of blue markers illustrates the superiority of DeepPrivacy2 compared to blurring and pixelation, as it indicates that the detector does not fail after anonymization. 
Pose estimation on pixelated images fails for 36 out of 38 images. 
In contrast to blurring, we observe a lower~EPE for DeepPrivacy2-anonymized images in all scenarios. 
A significant advantage of DeepPrivacy2 is visible for the "Visionlab machine" scenario. 
Here, DeepPrivacy2 achieves a PTK close to~100\% and an~EPE of less than~17~pixels for all images. 
Blurring, on the other hand, causes distance errors up to~238~pixels. 
The "Industrial machine" scenario shows one outlier with an EPE of~123~pixels. 
It is important to take the corresponding PTK into account. 
The outlier only considers four detected keypoints on this image. 

When visualizing the detected keypoints as in Fig.~\ref{fig:pose_estimation_rgb}, it becomes visible that the facial keypoints have a higher deviation than other keypoints. 
DeepPrivacy2 results in a different facial expression, which causes a different contour of the mouth. 
In blurred images, the jawline shows upwardly compressed keypoints, falsely suggesting a smaller face. 

\subsection{Action Recognition on Anonymized Videos}
The metrics for action recognition are calculated over the video dataset in dependence of a score threshold. 
Fig.~\ref{fig:action_dist} shows the top five actions (out of 60) and their distribution in the original and the anonymized videos. 

The human body detector fails for almost every pixelated frame.
Blurring and DeepPrivacy2 result in a similar action classification as in the original video. 
For action "carry/hold", DeepPrivacy2 is more accurate than blurring, which is evident from the deviation of more than~3,000 actions. 

\begin{figure}[h!]
    \centering
    \scalebox{0.95}{\input{plots/IPA_videos_accuracy}}
    \caption[Action accuracy of video dataset]{Action accuracy per anonymization type of~21,111 video frames in total as a function of the score threshold.}
    \label{fig:accuracy}
\end{figure}
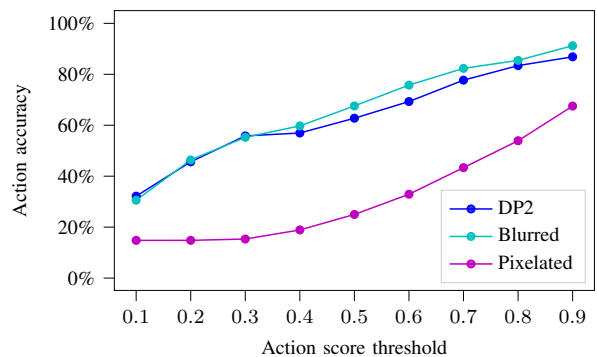

PTA in Fig.~\ref{fig:pta} specifies the percentage of original frames with detected actions in which actions are also detected after anonymization regardless of their correctness. 
For each score threshold, blurring has a~PTA of more than~87\%, which is lower than the~PTA of DeepPrivacy2. 
The~PTA of DeepPrivacy2 is almost~100\% until it drops to a value of~88\% at a score between 0.4 and 0.8. 
The lowest value is~75\% at a score of~0.9. 
It can be concluded that the anonymization by DeepPrivacy2 has the least impact on the detection performance of the action recognition framework. 

The accuracy in Fig.~\ref{fig:accuracy} indicates on how many anonymized video frames the action recognition recognizes the same actions as on the original frames. 
Blurring and DeepPrivacy2 show similar accuracy with a large margin to pixelation. 
The accuracy increases with the score threshold, as the number of desired actions per frame becomes smaller and thus the probability of hitting the exact action set~$\mathbb{X}_{true}$ becomes higher. 
In the original video, there are fewer recognized actions with a score of~0.9, so more empty sets $\mathbb{X}_{pred} = \mathbb{X}_{true} = \emptyset$ are compared and the accuracy increases. 
This explains why the accuracy curve for pixelated frames increases even though no actions are detected, as shown by the~PTA. 
In general, the accuracy for burring and DeepPrivacy2 is about~60\% on average.

%%%%%%%%%%%%%%%%%%%%%%%%%%%%%%%%%%%%%%%%%%%%%%%%%%%%%%%%%%%%%%%%%%%%%%%%%%%%%%%%
\section{CONCLUSION}

The experiments on the industrial setting show that blurring and pixelation lead to severe data degradation. 
Pixelation cause body detectors to fail and makes the data unusable for further applications such as pose estimation and action recognition. 
Blurring affects pose estimation so that less keypoints are detected and the position of facial keypoints has a high deviation. 
Full-body anonymization by DeepPrivacy2 shows a successful coverage of privacy-sensitive information. 
The applicability of a body detector for pose estimation and keypoint detection similar to the original data demonstrates the reuse of data and the production of realistic-looking artificial identities. 
In all scenarios, the images anonymized by DeepPrivacy2 have the lowest EPE compared to the other anonymization methods. 
Action recognition performs similar for blurring and DeepPrivacy2 with an accuracy around~60\%. 

It can be concluded that pixelation is not suitable as a feature-preserving anonymization method, since the anonymized data cannot be used for accurate pose estimation or action recognition. 
A low degree of pixelation reveals privacy-sensitive information. 
Blurring reduces the performance of pose estimation and action recognition, but the data is still usable. 
DeepPrivacy2 enables better pose estimation for body and face than blurring, but is not suitable for hand pose estimation for precision work. 
Furthermore, it affects the facial expression and is temporal inconsistent. 
The use of DeepPrivacy2 should be decided depending on the use case and shooting angle. 
In the given use case, facial expression is less relevant, but hand pose is more important. 
Nevertheless, DeepPrivacy2 is suitable for other industrial scenarios, e.g. for recording coarse work processes in a manufacturing facility, such as sorting large components or operating machines. 
In addition, generated artificial identities can serve as a data augmentation method to extend datasets for model training. 
For successful and smooth full-body anonymization, future work should address the problems of detailed body generation and temporal consistency in videos. 

%%%%%%%%%%%%%%%%%%%%%%%%%%%%%%%%%%%%%%%%%%%%%%%%%%%%%%%%%%%%%%%%%%%%%%%%%%%%%%%%

\end{document}

%% file: plots/IPA_images_EPE_score_50.tex
% This file was created with tikzplotlib v0.10.1.
\begin{tikzpicture}[font=\footnotesize, text_box/.style={rectangle, draw=black, text=black, thin, minimum height=5mm, minimum width=25mm}]

\definecolor{darkgray176}{RGB}{176,176,176}
\definecolor{darkturquoise0191191}{RGB}{0,191,191}
\definecolor{darkviolet1910191}{RGB}{191,0,191}
\definecolor{lightgray204}{RGB}{204,204,204}

\begin{groupplot}[group style={group size=2 by 2,  vertical sep=0.8cm, horizontal sep=0.5cm}]
\nextgroupplot[
height=4.5cm,
legend cell align={left},
legend style={
  fill opacity=0.8,
  draw opacity=1,
  text opacity=1,
  at={(0.03,0.05)},
  anchor=south west,
  draw=lightgray204
},
tick align=outside,
tick pos=left,
width=6.5cm,
x grid style={darkgray176},
xmin=1.5, xmax=13.5,
xtick style={color=black},
xtick={2,4,6,8,10,12},
xticklabels={2,4,6,8,10,12},
y grid style={darkgray176},
ymin=-0.0164, ymax=1.0484,
ytick style={color=black},
ytick={-0.2,0,0.2,0.4,0.6,0.8,1,1.2},
yticklabels={\ensuremath{-}20\%,0\%,20\%,40\%,60\%,80\%,100\%,120\%}
]
\addplot [draw=blue, fill=blue, mark=*, only marks]
table{%
x  y
5.3654 0.9587
4.7262 1
12.3945 1
4.546 1
7.414 1
6.1264 0.88
4.1477 1
9.0986 1
8.6072 0.9902
3.8983 1
};
\addlegendentry{DP2}
\addplot [draw=darkturquoise0191191, fill=darkturquoise0191191, mark=*, only marks]
table{%
x  y
8.9079 0.8595
6.568 1
8.9704 1
5.138 1
10.0345 1
6.7311 0.992
5.4847 1
7.872 1
6.1042 1
8.8399 1
};
\addlegendentry{Blurred}
\addplot [draw=darkviolet1910191, fill=darkviolet1910191, mark=*, only marks]
table{%
x  y
11.4243 0.936
5.9261 0.9902
};
\addlegendentry{Pixelated}

\nextgroupplot[
height=4.5cm,
scaled y ticks=manual:{}{\pgfmathparse{#1}},
tick align=outside,
tick pos=left,
width=6.5cm,
x grid style={darkgray176},
xmin=1.5, xmax=13.5,
xtick style={color=black},
xtick={2,4,6,8,10,12},
xticklabels={2,4,6,8,10,12},
y grid style={darkgray176},
ymin=-0.0164, ymax=1.0484,
ytick style={color=black},
ytick={-0.2,0,0.2,0.4,0.6,0.8,1,1.2},
yticklabels={}
]
\addplot [draw=blue, fill=blue, mark=*, only marks]
table{%
x  y
3.336 0.932
3.7099 0.8145
6.2208 1
5.1711 1
4.0366 0.9706
7.8421 1
7.3639 1
4.8247 0.992
3.9457 0.8151
};
\addplot [draw=darkturquoise0191191, fill=darkturquoise0191191, mark=*, only marks]
table{%
x  y
11.3643 0.9126
6.9432 0.9435
5.2127 0.7841
6.703 1
11.1745 0.9902
12.9593 1
12.2357 0.992
8.1486 1
10.009 0.8655
};

\nextgroupplot[
height=4.5cm,
tick align=outside,
tick pos=left,
width=6.5cm,
x grid style={darkgray176},
xmin=-5, xmax=255,
xtick style={color=black},
xtick={0,50,100,150,200,250},
xticklabels={0,50,100,150,200,250},
y grid style={darkgray176},
ymin=-0.0164, ymax=1.0484,
ytick style={color=black},
ytick={-0.2,0,0.2,0.4,0.6,0.8,1,1.2},
yticklabels={\ensuremath{-}20\%,0\%,20\%,40\%,60\%,80\%,100\%,120\%}
]
\addplot [draw=blue, fill=blue, mark=*, only marks]
table{%
x  y
4.7244 1
5.1753 1
6.2845 0.9914
8.4513 0.9608
16.9985 0.6
8.0048 1
8.4268 0.8571
12.1528 1
};
\addplot [draw=darkturquoise0191191, fill=darkturquoise0191191, mark=*, only marks]
table{%
x  y
27.3139 0.1557
237.5684 0.098
6.4341 0.5556
130.3304 0.06
};

\nextgroupplot[
height=4.5cm,
scaled y ticks=manual:{}{\pgfmathparse{#1}},
tick align=outside,
tick pos=left,
width=6.5cm,
x grid style={darkgray176},
xmin=-1.954765, xmax=129.232665,
xtick style={color=black},
xtick={0,20,40,60,80,100,120},
xticklabels={0,20,40,60,80,100,120},
y grid style={darkgray176},
ymin=-0.0164, ymax=1.0484,
ytick style={color=black},
ytick={-0.2,0,0.2,0.4,0.6,0.8,1,1.2},
yticklabels={}
]
\addplot [draw=blue, fill=blue, mark=*, only marks]
table{%
x  y
5.7059 0.9837
7.6692 0.8364
123.2696 0.032
5.5937 0.2791
7.2604 0.27
4.0083 0.3708
5.5441 0.9919
13.9714 1
7.8397 0.7857
39.2335 0.3204
};
\addplot [draw=darkturquoise0191191, fill=darkturquoise0191191, mark=*, only marks]
table{%
x  y
19.0453 0.2033
9.8501 0.9182
7.7158 0.5565
};
\end{groupplot}

\draw ({$(current bounding box.south west)!0.54!(current bounding box.south east)$}|-{$(current bounding box.south west)!-0.05!(current bounding box.north west)$}) node[
  scale=1.2,
  anchor=base,
  text=black,
  rotate=0.0
]{EPE in pixel};
\draw ({$(current bounding box.south west)!-0.01!(current bounding box.south east)$}|-{$(current bounding box.south west)!0.56!(current bounding box.north west)$}) node[
  scale=1.2,
  anchor=base,
  text=black,
  rotate=90.0
]{PTK};
\node[text_box, align=center] (text_box_1) at (3.5,0.5)    {Office};
\node[text_box, align=center] (text_box_2) at (8.9,0.5)   {Visionlab assembly};
\node[text_box, align=center] (text_box_3) at (3.5,-1.28)  {Visionlab machine};
\node[text_box, align=center] (text_box_4) at (8.9,-1.28) {Industrial machine};
\end{tikzpicture}

%% file: plots/IPA_videos_action_distribution.tex
% This file was created with tikzplotlib v0.10.1.
\begin{tikzpicture}[font=\footnotesize]

\definecolor{darkgray176}{RGB}{176,176,176}
\definecolor{darkorange25512714}{RGB}{255,127,14}
\definecolor{darkturquoise0191191}{RGB}{0,191,191}
\definecolor{darkviolet1910191}{RGB}{191,0,191}
\definecolor{lightgray204}{RGB}{204,204,204}

\begin{axis}[
height=5.5cm,
width=8.3cm,
legend cell align={left},
legend style={fill opacity=0.8, draw opacity=1, text opacity=1, draw=lightgray204},
tick align=outside,
tick pos=left,
x grid style={darkgray176},
xmin=-0.4875, xmax=7.4875,
xtick style={color=black},
xtick={0.375,1.625,2.875,4.125,5.375,6.625},
xticklabels={Bend/bow,Stand,Walk,Carry/hold,Touch,Watch},
y grid style={darkgray176},
xlabel={Selected actions},
ylabel={Occurrence in frames},
ymin=0, ymax=17000,
scaled ticks=false, 
tick label style={/pgf/number format/fixed},
ytick style={color=black},
ytick={2000,4000,6000,8000,10000,12000,14000,16000},
]
\draw[draw=none,fill=darkorange25512714] (axis cs:-0.125,0) rectangle (axis cs:0.125,4302);
\addlegendimage{ybar,ybar legend,draw=none,fill=darkorange25512714}
\addlegendentry{Original}

\draw[draw=none,fill=darkorange25512714] (axis cs:1.125,0) rectangle (axis cs:1.375,15377);
\draw[draw=none,fill=darkorange25512714] (axis cs:2.375,0) rectangle (axis cs:2.625,2251);
\draw[draw=none,fill=darkorange25512714] (axis cs:3.625,0) rectangle (axis cs:3.875,16057);
\draw[draw=none,fill=darkorange25512714] (axis cs:4.875,0) rectangle (axis cs:5.125,4210);
\draw[draw=none,fill=darkorange25512714] (axis cs:6.125,0) rectangle (axis cs:6.375,2029);
\draw[draw=none,fill=blue] (axis cs:0.125,0) rectangle (axis cs:0.375,3996);
\addlegendimage{ybar,ybar legend,draw=none,fill=blue}
\addlegendentry{DP2}

\draw[draw=none,fill=blue] (axis cs:1.375,0) rectangle (axis cs:1.625,15252);
\draw[draw=none,fill=blue] (axis cs:2.625,0) rectangle (axis cs:2.875,3164);
\draw[draw=none,fill=blue] (axis cs:3.875,0) rectangle (axis cs:4.125,16301);
\draw[draw=none,fill=blue] (axis cs:5.125,0) rectangle (axis cs:5.375,2854);
\draw[draw=none,fill=blue] (axis cs:6.375,0) rectangle (axis cs:6.625,1977);
\draw[draw=none,fill=darkturquoise0191191] (axis cs:0.375,0) rectangle (axis cs:0.625,3495);
\addlegendimage{ybar,ybar legend,draw=none,fill=darkturquoise0191191}
\addlegendentry{Blurred}

\draw[draw=none,fill=darkturquoise0191191] (axis cs:1.625,0) rectangle (axis cs:1.875,15771);
\draw[draw=none,fill=darkturquoise0191191] (axis cs:2.875,0) rectangle (axis cs:3.125,1679);
\draw[draw=none,fill=darkturquoise0191191] (axis cs:4.125,0) rectangle (axis cs:4.375,12822);
\draw[draw=none,fill=darkturquoise0191191] (axis cs:5.375,0) rectangle (axis cs:5.625,2985);
\draw[draw=none,fill=darkturquoise0191191] (axis cs:6.625,0) rectangle (axis cs:6.875,1349);
\draw[draw=none,fill=darkviolet1910191] (axis cs:0.625,0) rectangle (axis cs:0.875,100);
\addlegendimage{ybar,ybar legend,draw=none,fill=darkviolet1910191}
\addlegendentry{Pixelated}

\draw[draw=none,fill=darkviolet1910191] (axis cs:1.875,0) rectangle (axis cs:2.125,100);
\draw[draw=none,fill=darkviolet1910191] (axis cs:3.125,0) rectangle (axis cs:3.375,108);
\draw[draw=none,fill=darkviolet1910191] (axis cs:4.375,0) rectangle (axis cs:4.625,101);
\draw[draw=none,fill=darkviolet1910191] (axis cs:5.625,0) rectangle (axis cs:5.875,100);
\draw[draw=none,fill=darkviolet1910191] (axis cs:6.875,0) rectangle (axis cs:7.125,107);

\end{axis}
\end{tikzpicture}

%% file: plots/IPA_videos_PTA.tex
% This file was created with tikzplotlib v0.10.1.
\begin{tikzpicture}[font=\footnotesize]

\definecolor{darkgray176}{RGB}{176,176,176}
\definecolor{darkturquoise0191191}{RGB}{0,191,191}
\definecolor{darkviolet1910191}{RGB}{191,0,191}
\definecolor{lightgray204}{RGB}{204,204,204}

\begin{axis}[
height=5.5cm,
legend cell align={left},
legend style={
  fill opacity=0.8,
  draw opacity=1,
  text opacity=1,
  at={(0.98,0.36)},
  anchor=east,
  draw=lightgray204
},
tick align=outside,
tick pos=left,
width=8.3cm,
x grid style={darkgray176},
xlabel={Action score threshold},
xmin=0.06, xmax=0.94,
xtick={0.1,0.2,0.3,0.4,0.5,0.6,0.7,0.8,0.9},
xtick style={color=black},
y grid style={darkgray176},
ylabel={PTA},
ymin=-0.0484, ymax=1.05,
ytick style={color=black},
ytick={0,0.2,0.4,0.6,0.8,1,1.2},
yticklabels={0\%,20\%,40\%,60\%,80\%,100\%,120\%}
]
\addplot [semithick, blue, mark=*, mark size=1.5, mark options={solid}]
table {%
0.1 0.9968
0.2 0.9966
0.3 0.9886
0.4 0.9731
0.5 0.9393
0.6 0.9089
0.7 0.911
0.8 0.8886
0.9 0.7529
};
\addlegendentry{DP2}
\addplot [semithick, darkturquoise0191191, mark=*, mark size=1.5, mark options={solid}]
table {%
0.1 0.9363
0.2 0.9362
0.3 0.9342
0.4 0.9193
0.5 0.9
0.6 0.8737
0.7 0.8818
0.8 0.8836
0.9 0.8773
};
\addlegendentry{Blurred}
\addplot [semithick, darkviolet1910191, mark=*, mark size=1.5, mark options={solid}]
table {%
0.1 0.0004
0.2 0.0004
0.3 0.0004
0.4 0.0005
0.5 0.0005
0.6 0.0006
0.7 0.0007
0.8 0.0002
0.9 0
};
\addlegendentry{Pixelated}
\end{axis}

\end{tikzpicture}

%% file: plots/IPA_videos_accuracy.tex
% This file was created with tikzplotlib v0.10.1.
\begin{tikzpicture}[font=\footnotesize]

\definecolor{darkgray176}{RGB}{176,176,176}
\definecolor{darkturquoise0191191}{RGB}{0,191,191}
\definecolor{darkviolet1910191}{RGB}{191,0,191}
\definecolor{lightgray204}{RGB}{204,204,204}

\begin{axis}[
height=5.5cm,
legend cell align={left},
legend style={
  fill opacity=0.8,
  draw opacity=1,
  text opacity=1,
  at={(0.68,0.36)},
  anchor=north west,
  draw=lightgray204
},
tick align=outside,
tick pos=left,
width=8.3cm,
x grid style={darkgray176},
xlabel={Action score threshold},
xmin=0.06, xmax=0.94,
xtick={0.1,0.2,0.3,0.4,0.5,0.6,0.7,0.8,0.9},
xtick style={color=black},
y grid style={darkgray176},
ylabel={Action accuracy},
ymin=-0.0484, ymax=1.05,
ytick style={color=black},
ytick={0,0.2,0.4,0.6,0.8,1,1.2},
yticklabels={0\%,20\%,40\%,60\%,80\%,100\%,120\%}
]
\addplot [semithick, blue, mark=*, mark size=1.5, mark options={solid}]
table {%
0.1 0.3216
0.2 0.4565
0.3 0.5582
0.4 0.5695
0.5 0.6276
0.6 0.693
0.7 0.777
0.8 0.8341
0.9 0.8686
};
\addlegendentry{DP2}
\addplot [semithick, darkturquoise0191191, mark=*, mark size=1.5, mark options={solid}]
table {%
0.1 0.3058
0.2 0.4638
0.3 0.5526
0.4 0.5974
0.5 0.6758
0.6 0.7575
0.7 0.8233
0.8 0.8546
0.9 0.9121
};
\addlegendentry{Blurred}
\addplot [semithick, darkviolet1910191, mark=*, mark size=1.5, mark options={solid}]
table {%
0.1 0.148
0.2 0.1481
0.3 0.1533
0.4 0.189
0.5 0.2498
0.6 0.3291
0.7 0.4335
0.8 0.5391
0.9 0.6753
};
\addlegendentry{Pixelated}
\end{axis}

\end{tikzpicture}